\documentclass[authoryear,12pt]{elsarticle}
\usepackage{geometry}
\usepackage[utf8]{inputenc}
\usepackage{newunicodechar}
\newunicodechar{−}{\ensuremath{-}}
\usepackage{algpseudocode} 
\usepackage{amsmath}
\usepackage{algpseudocodex} 
\usepackage{algorithmicx,algorithm}
\usepackage{setspace} 
\usepackage{graphicx} 
\usepackage{epstopdf}
\usepackage{eurosym} 
\usepackage{amssymb}
\usepackage{gensymb}
\usepackage{subfigure}
\usepackage[authoryear]{natbib}
\usepackage{mathrsfs} 
\usepackage[colorlinks=true,linkcolor=blue,citecolor=blue]{hyperref}%
\usepackage[a-1b]{pdfx} 

\usepackage{tabulary}
\usepackage{url}
\usepackage{booktabs}    
\usepackage{tabularx}    
\usepackage{makecell}    
\usepackage{siunitx}     
\usepackage{array}       
\newcommand{\coords}[3]{\makecell{(#1)\\(#2)\\\textbf{#3}}}
\newcommand{\coordsIII}[3]{\makecell{\textbf{#1}\\(#2)\\(#3)}}
\newcommand{\coordsI}[2]{\makecell[l]{(#1)\\(#2)}}
\usepackage{makecell}       
\newcommand{\coordsII}[2]{\makecell{(#1)\\\textbf{#2}}}
\usepackage{tabularx,array}
\newcolumntype{W}[1]{>{\hsize=#1\hsize\arraybackslash}X} 
\usepackage{caption}
\usepackage{subcaption}   
\usepackage{booktabs,siunitx,makecell,ragged2e}
\usepackage[english]{babel}
\usepackage{lmodern} 
\usepackage[T1]{fontenc}
\usepackage[babel=true]{microtype} 
\usepackage{booktabs}
\usepackage{xcolor}
\usepackage{gensymb}
\usepackage{multirow}

\begin{document}

\begin{frontmatter}

\title{Optimal placement of wind farms via quantile constraint learning}

\author[inst1]{Wenxiu Feng\corref{cor1}}
\ead{100475987@alumnos.uc3m.es}
\author[inst3]{Antonio Alcántara}
\ead{anmata@dtu.dk}
\author[inst1,inst2]{Carlos Ruiz}
\ead{caruizm@est-econ.uc3m.es}
\cortext[cor1]{Corresponding author}

 \affiliation[inst1]{organization={Department of Statistics, University Carlos III of Madrid},
             country={Spain}}

 \affiliation[inst2]{organization={UC3M-BS Institute for Financial Big Data (IFiBiD), University Carlos III of Madrid},
             country={Spain}}

 \affiliation[inst3]{organization={Department of Wind and Energy Systems, Technical University of Denmark (DTU)},
             country={Denmark}}


 \begin{abstract}
Wind farm placement arranges the size and the location of multiple wind farms within a given region. The power output is highly related to the wind speed on spatial and temporal levels, which can be modeled by advanced data-driven approaches. To this end, we use a probabilistic neural network as a surrogate that accounts for the spatiotemporal correlations of wind speed. This neural network uses ReLU activation functions so that it can be reformulated as mixed-integer linear set of constraints (constraint learning). We embed these constraints into the placement decision problem, formulated as a two-stage stochastic optimization problem. Specifically, conditional quantiles of the total electricity production are regarded as recursive decisions in the second stage. We use real high-resolution regional data from a northern region in Spain. We validate that the constraint learning approach outperforms the classical bilinear interpolation method. Numerical experiments are implemented on risk-averse investors. The results indicate that risk-averse investors concentrate on dominant sites with strong wind, while exhibiting spatial diversification and sensitive capacity spread in non-dominant sites. Furthermore, we show that if we introduce transmission line costs in the problem, risk-averse investors favor locations closer to the substations. On the contrary, risk-neutral investors are willing to move to further locations to achieve higher expected profits. Our results conclude that the proposed novel approach is able to tackle a portfolio of regional wind farm placements and further provide guidance for risk-averse investors.

 \end{abstract}

\end{frontmatter}

\section{Introduction}

The world is undertaking an important process towards the de-carbonization and sustainability of all types of industries. As part of the green and renewable energy system, wind-based energy production has been widely deployed for its competitive cost, mature technology and extended geographical and meteorological compatibility. During the last 20 years, the amount of wind power capacity has experienced an exponential growth, and is expected to reach 5000 TWh electricity generation, almost 15\% of global electricity generation, by 2030 \citep{IEA:Renewables2024}. 

In recent years, the investment and expansion of onshore wind have been promoted in Europe, the United States, and Southeast Asia \citep{IEA:Renewables2024}. During the rapid expansion of onshore wind power, wind farm placement raised great concerns. Prospective investors prefer to establish wind turbines at sites with high and stable wind generation profiles. However, the intermittent nature of wind poses great challenges to the stability and reliability of the power generation. The portfolio placement of onshore wind farms, which locates several wind farms at distant sites, is proposed to smooth fluctuations of wind energy generation \citep{reichenberg2014dampening, poulsen2022optimization}. Current spatiotemporal techniques are implemented on finite sites but with high complexity. The difficulties increase when the optimal wind farm sites are explored. In this regard, deriving aggregated wind production profiles across continuous sites within one region is a challenging task.

Another important aspect is that wind farms need to deliver their electricity to a substation via transmission lines. Electricity delivery cost, known as the cost of transmission lines/grid connection, is one of the key costs that affect the profits of investors. Investors aim to establish the most economically favorable connection sites, which is hard to implement in reality \citep{BWE_Netzanschlussoptimierung}. Therefore, the optimal wind farm must balance the hourly income from electricity generation with the long-term investment cost of the transmission lines. 

We focus on the problem faced by a public or private wind investor, that wants to locate more than one wind farm within a given region, with the goal of maximizing the total profit in a target year. The investor has a limited budget of economical resources so that it seeks to identify the most profitable locations and wind farm sizes. However, wind speed distributions are temporally and spatially correlated in the region, which conditions the aggregate wind power generation of the portfolio. On the one hand, two wind farms with close and highly spatially correlated locations produce, within a year, a large amount of energy. However, they also entail a large number of hours with null total electricity production, since low wind conditions affect both installations simultaneously. On the other hand, two wind farms located at more distant and uncorrelated locations may not render such a large amount of total production within a given year. However, they can better complement each other to reduce the number of hours with a null total production. These aspects are especially important for risk-averse investors, who are not solely interested in large expected profits but also in reducing the probability of low profit outcomes.

To this end, we employ a data-driven approach to approximate complicated spatial wind dynamics interactions as a surrogate CL model, the incremental quantile neural network (IQNN), which is learned from spatial granular wind data. In particular, this model characterizes the functional relationship between Total Production \textit{vs} Wind Farms Locations and Sizes, and can be embedded as a set of constraints into the placement optimization problem.

The proposed wind farm placement problem models the decision of risk-averse investors. It is formulated as two-stage stochastic programming, which decides siting and sizing in the first stage, and regards conditional wind power quantiles as second-stage scenarios. Moreover, we incorporate and investigate the necessary connection cost to the substation. Co-optimizing the wind revenue and cost provides economical insights for risk-averse investors.

The main contributions of this work are fourfold:
\begin{enumerate}
    \item We propose a fully data-driven optimization approach for the sitting and sizing of several wind farms within a given region. 
    \item The complex spatio-temporal dynamics of wind power generation are approximated by a probabilistic incremental quantile neural network (IQNN). The IQNN provides location-dependent wind generation quantiles, outperforming standard interpolation methods.
    \item The IQNN is embedded, via constraint learning techniques, within the sitting and sizing stochastic optimization problem, which is efficient to solve with available solvers. This allows to explicitly model the impact of continuous locational decisions in power generation under uncertainty, differentiating from most of the literature that assume a finite set of candidate sites \citep{cetinay2017optimal,poulsen2022optimization}. The model incorporates connection costs and risk aversion. 
    \item We employ real data from ERA5-Land climate reanalysis, where we investigate the role of the region characteristics, the investor’s risk-aversion level, and the impact of the transmission line investment costs on the optimal sitting and sizing strategy.
\end{enumerate}

The remainder of this paper is organized as follows. Section \ref{sec:litrev} provides a summary of the literature review related to the wind farm placement problem and the intersection of constraint learning and energy applications. Section \ref{sec:methodology} gives an overview of the surrogate predictive model and presents the two-stage stochastic model used in the optimization. Section \ref{CR} describes sample instances used for the surrogate model and compares its performance with the baseline interpolation method. Then, practical case studies of risk-averse investors are investigated. Finally, Section \ref{Conclusion} summarizes the conclusions and discusses possible extensions.

\section{Literature review}
\label{sec:litrev}

\subsection{Wind farm siting}

Research on wind farms from a location perspective follows two main direction: macro-scale wind farm siting \citep{cetinay2017optimal, poulsen2022optimization} and micro-scale turbine layout and cable routing optimization \citep{fischetti2019machine, pedersen2024efficient}. Our study addresses the first. Investors seek large-scale wind farms on windy sites, so they need a reliable estimation of wind speed distribution for each site. The wind speed distribution at a specific site is generally determined by time series methods \citep{morales2010methodology,katikas2021stochastic,zhang2025weather} and statistical analysis methods \citep{dhople2012framework,cetinay2017optimal,dos2024brazilian}. Time series methods require enormous long-term historical data. Statistical approaches are used when historical data is scarce. \cite{poulsen2022optimization} use the Weibull distribution, further to assess the wind characteristics at a geographical location and determine the feasible sites for the wind farms. To ensure reliability and stability, risk analysis concerning extreme or maximum wind speeds is conducted using the extreme value distribution \citep{xiao2006probability,kang2015determination}. The Gamma distribution is also widely used in modeling wind speed \citep{aries2018deep}. In addition to the parametric distribution models, non‑parametric techniques such as kernel density estimation \citep{han2019kernel} and the maximum‑entropy principle \citep{chellali2012comparison} learn the distribution directly from the data.

The studies above focus on selecting a single site. For optimal placement of wind farm portfolios, the literature is split into two lines. One examines the smoothing effect, analyzing minutely or hourly time series from each site to reduce operational‑horizon fluctuations \citep{hasche2010general,dalala2013design,yang2019investigating,poulsen2020spectral,poulsen2022optimization}; the other aggregates the long‑term distributions of individual sites to estimate annual revenue or risk. Our study focuses on the latter. Previous research on the smoothing effect that reduces power fluctuations through correlation analysis \citep{yang2019investigating}, step change analysis \citep{dalala2013design}, visualization of duration curves \citep{hasche2010general}, and frequency domain approaches \citep{poulsen2020spectral}. \cite{poulsen2022optimization} present that hourly modeled wind turbine power output series, analyzed with power spectral density, hourly step change function, and duration curves, are used to design wind farm portfolios with lower output fluctuations. For distribution‑based planning, mixtures of Weibull distributions are fitted at each site, and inverse‑distance weighting is used to create continuous wind maps \citep{dos2024brazilian}. A joint distribution for three wind farms has been obtained by coupling Principal Component Analysis‑compressed data with an R‑vine copula \citep{goh2022new}. The Gaussian copulas method is employed for high‑dimensional probabilistic forecasting, capturing the spatial and temporal dependence of energy production across wind farms and solar plants \citep{zhang2025weather}. 

Previous studies demonstrate that characterizing wind generation is complex even for a single site. The complexity grows when the portfolio of wind farm placement is involved. Most studies evaluate wind profiles only at given candidate sites, and few studies consider long-distance grid connection costs. The goal of this study is to provide an effective approach for joint siting and sizing of onshore wind farms over a large region. Wind farm investors are assumed to be risk-averse. The problem investigates how optimal placements respond to aggregate wind profiles and transmission line costs.

\subsection{Constraint learning applications in energy}

Constraint learning (CL) is used in this work to model the complex spatial dependency of wind generation. CL is increasingly recognized as a bridge between data-driven modeling and mathematical optimization. In CL, predictive models (most commonly neural networks, decision trees, or ensemble methods) approximate otherwise intractable or unknown relationships between decision variables $x$ and responses $y$ to be then embedded as algebraic constraints inside mixed-integer linear or nonlinear programs. Early work demonstrated that once a surrogate model is reformulated as linear or mixed-integer linear constraints, classical solvers can deal with decision problems that were previously computationally prohibitive \citep{maragno2023mixed, fajemisin2024optimization}.

Energy systems have provided a fertile ground for CL experimentation. Surrogates have cut simulation time in biodiesel reactors \citep{fahmi2012process}, screened secure operating points in transmission grids \citep{cremer2018data, halilbavsic2018data}, and delivered fast voltage-control policies \citep{chen2020input}. More recently, researchers embedded ReLU neural networks to speed up iterative AC-power-flow feasibility checks \citep{zhou2024accelerating}, and coupled quantile or distributional surrogates with stochastic programs to capture risk in electricity bidding \citep{alcantara2024optimal, alcantara2025quantile}. Collectively, these contributions confirm that CL can handle high-dimensional decision spaces, sharply nonlinear responses, and uncertainty, which also characterize the wind farm placement problem.

\section{Methodology} \label{sec:methodology}

\subsection{Surrogate model: IQNN for wind power quantiles}\label{sec_3_1}

Mathematically, the general framework of an optimization problem with embedded surrogate model is as follows:
\begin{subequations}\label{eq:opt_CL}
\begin{align}
    \min_{x \in \mathbb{R}^{n_0}, y \in \mathbb{R}^k} & \quad f(x, z, y) \\
    \text{s.t.} & \quad g(x, z, y) \leq 0, \\
                & \quad y = \hat{h}_{\theta}(x, z) \label{eq:surrog}
\end{align}
\end{subequations}
\noindent where $f(x, z, y)$ is the objective function, $g(x, z, y)$ represents explicit constraints, $\hat{h}_{\theta}(x, z)$ is the learned surrogate model with parameters $\theta$, and $z$ defines possible contextual information. The surrogate model (\ref{eq:surrog}) captures the relationship between the response variable $y$ and $x$, both considered as optimization variables in \eqref{eq:opt_CL}. In our setting, $x$ would represent the size and location, and $y$ the power output of the wind farms.

\begin{figure}[!ht] 
\centering 
\includegraphics[width=1\textwidth]{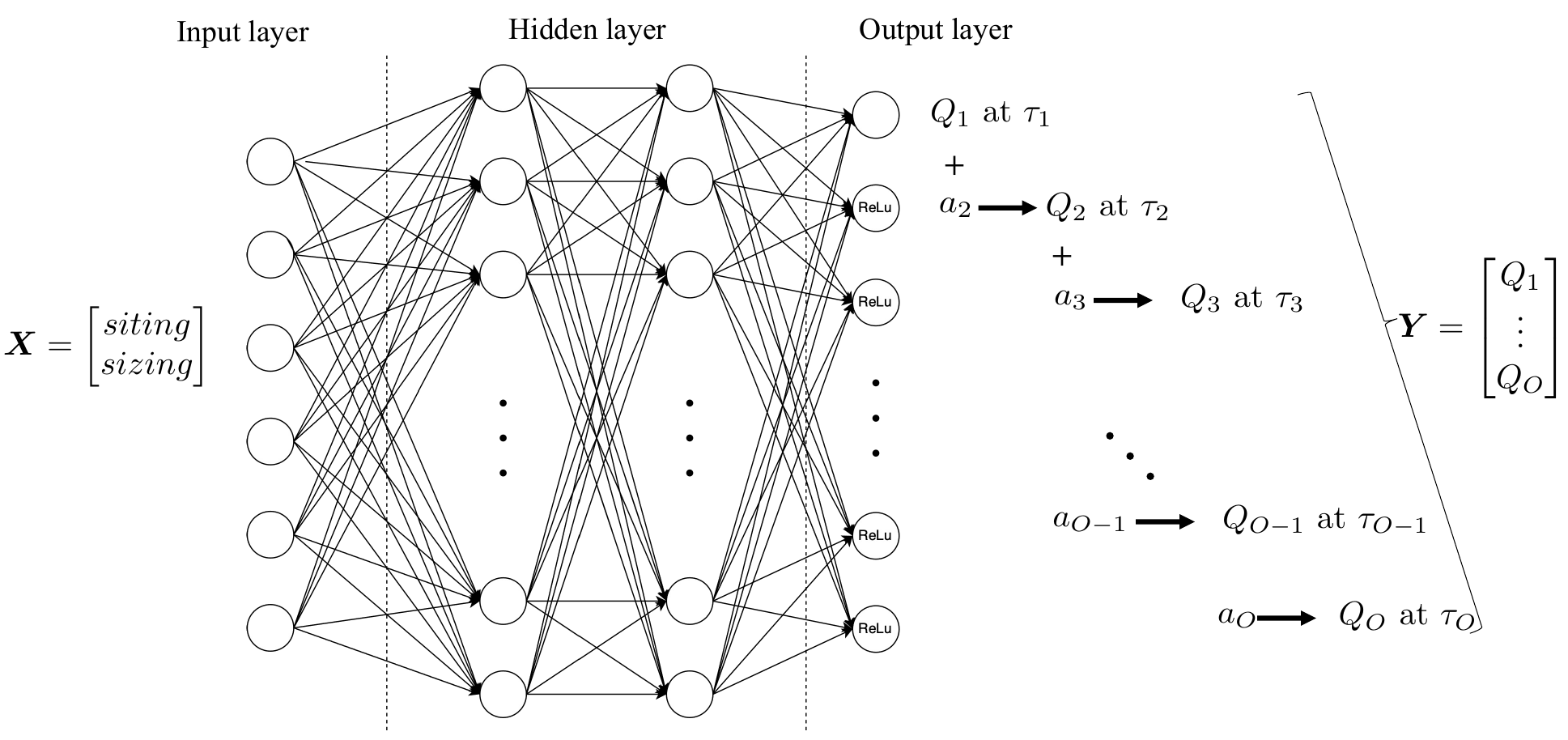} 
\caption{Incremental Quantile Neural Network (IQNN) for wind power quantiles. Figure and caption adapted from \cite{alcantara2025quantile}.} %
\label{Fig.IQNN} %
\end{figure}

In particular, IQNN \citep{alcantara2025quantile} is employed as our surrogate model $\hat{h}$, which maps siting and sizing variables to multiple quantiles of the total energy generation. Specifically, Figure \ref{Fig.IQNN} shows the structure of the IQNN with one input layer, multiple hidden layers, and one output layer. It differentiates from a standard neural network in the output layer, which produces multiple quantile values instead of one single output. Outputs of conditional quantiles are able to estimate and capture the wind generation distribution.

The IQNN is defined as follows, mathematically. We set decision variables of sites and sizes $\textbf{X} =[siting, sizing]^{\top}$ as inputs. Each neuron contains calculations of weighted summation in equation \eqref{summation} and activation in equation \eqref{activation}. 
\begin{subequations}
\begin{align}
    \label{summation}& \quad \textbf{z}^{l}=\textbf{w}^{l}\textbf{a}^{l-1}+\textbf{b}^{l}\ \\
    \label{activation}& \quad \textbf{a}^{l} = g^{l}(\textbf{z}^{l})
\end{align}
\end{subequations}
where $a^{0}=\textbf{X}$ are the input, $a_{j=2,\dots,O}^{L+1}=[Q_{2}-Q_{1},\dots,Q_{O}-Q_{O-1}]^{\top}$ are the output increments of wind generation quantiles, and $g$ denotes activation function. Weights matrix $\textbf{w}$ and bias vector $\textbf{b}$ are optimized by minimizing the multi-quantile loss \citep{hao2007quantile,alcantara2025quantile} in \eqref{multi-quantileLoss}. 
\begin{equation}\label{multi-quantileLoss}
    \quad (\textbf{w},\textbf{b})^\star=
    \arg\min_{\textbf{w},\textbf{b}}\frac{1}{N K}\sum_{i=1}^{N}\sum_{j=1}^{K}\bigl(\tau_j\max(\epsilon_{i,j},0)+(1-\tau_j)\,\max(-\epsilon_{i,j},0)\bigr)    
\end{equation}
where $\epsilon_{i,j} = y_i - \hat{Q}_j{(\textbf{X}_i)},\ i=1,\dots,N, j=1,\dots,K$ is the estimated residual on sample $i$ at quantile level $\tau_j$.

More specifically, each hidden layer uses the Rectified Linear Unit (ReLU) activation function. To embed the IQNN into the optimization model, ReLU activation functions $a=\max\{0,z\}$, $z \in \left [M_{\underline{z}}, M_{\overline{z}} \right ]$, are rewritten as a set of piecewise mixed-integer linear constraints \eqref{LinearbigM} \citep{tjeng2017evaluating,bunel2018unified,alcantara2025quantile}. 
\begin{equation}\label{LinearbigM}
       a \geq 0; \quad a \geq z; \quad a  \leq z-M_{\underline{z}}(1-\delta); \quad a \leq M_{\overline{z}}\delta
\end{equation}
where $\delta \in\{0,1\}$ is incorporated as a binary variable to model that the ReLU function is active ($\delta=1$) with $a=z$  or inactive ($\delta=0$) with $a=0$.

\subsection{Optimization model with embedded IQNN}

Each wind farm location and number of turbines are determined by a trade-off between cost and revenue. The objective function considers the transmission connection cost and the Conditional Value at Risk (CVaR) of the total wind generation. The main notation used in the model formulation \eqref{P1} is introduced in the following as a reference.

\noindent \textbf{Indices:}
\begin{itemize}
  \item[$s$] Index for selected wind farm sites, where $s\in \{1,\dots,S\}$ \\
 \vspace{-0.85cm}
  \item[$f$] Index for load center/ substation, where $f\in \{1,\dots,F\}$\\
 \vspace{-0.85cm}
  \item[$o$] Index for quantiles of total wind energy production, where $o\in \{1,...,n^{L+1}=O\}$\\
 \vspace{-0.85cm}
  \item[$l$] Index for layers of the IQNN, where $l\in \{1,...,L+1\}$\\
 \vspace{-0.85cm}
  \item[$j$] Index for neurons at layer $l$ of the IQNN, where $j \in \{1,...,n^{l}\}$
 \end{itemize}
\noindent \textbf{Parameters:}
\begin{itemize}
  \item[$c^{L}$] The unit cost of transmission line\\
 \vspace{-0.85cm}
  \item[$\lambda^{\rm{PPA}}$] The power purchase agreement price for electricity\\
 \vspace{-0.85cm}
  \item[$\alpha$] The significance level used to model risk-aversion\\
 \vspace{-0.85cm}
   \item[$\tau_{o}$] The quantile level for the $o$th quantile\\
 \vspace{-0.85cm}
  \item[$\overline{N}$] The fixed total number of wind turbines\\
 \vspace{-0.85cm}
  \item[$\textbf{X}_{\min}, \textbf{X}_{\max}$] The boundary of siting and sizing decision variables at the input layer \\
 \vspace{-0.85cm}
  \item[$\textbf{Y}_{\min}, \textbf{Y}_{\max}$] The boundary of wind power quantiles at the output layer\\
    \vspace{-0.85cm}
   \item[$n^{l}$] The number of neurons at layer $l$ in the trained IQNN\\
   \vspace{-0.85cm}
   \item[$w^{l}_{ij}$] The weight term between neuron $i$ and $j$ of layer $l$ in the trained IQNN\\
 \vspace{-0.85cm}
   \item[$b^{l}_{j}$] The bias term at neuron $j$ of  layer $l$ in the trained IQNN\\
 \vspace{-0.85cm}
  \item[$M^{-,l}_{j},M^{+,l}_{j}$] Large constant value (Big M) for $j \in \{1,...,n^{l}\}, l \in \{1,...,L\}$
 \end{itemize}
 \noindent \textbf{Variables:}

 \begin{itemize}
 \item[$x_s$] location in terms of coordinates (latitude,longitude) of wind farm site $s\in \{1,\dots,S\}$\\
 \vspace{-0.85cm}
   \item[$n_{s}$] The number of wind turbines at wind farm site $s\in \{1,\dots,S\}$\\
 \vspace{-0.85cm}
  \item[$\textbf{X}$] The siting and sizing decision variables $\textbf{X} =\left[x_{1},\dots, x_{S}, n_{1},\dots, n_{S}\right]^T$ at the input layer\\
\vspace{-0.85cm}
   \item[$u_{sf}$] The binary variable that represents whether the wind farm at site $s$ is connected to substation $f$.\\
 \vspace{-0.85cm}
   \item[$d_{sf}$] The distance between the wind farm site $s$ and substation $f$\\
 \vspace{-0.85cm}
   \item[$q_{o}$] The quantile value for the $o$th wind energy output\\
 \vspace{-0.85cm}
   \item[$Q_{o}$] The rescaled quantile value $o$th wind energy output\\
 \vspace{-0.85cm}
   \item[$\textbf{Y}$] The wind power quantile values $\textbf{Y}=[Q_{1},\dots,Q_{O}]^{T}$ at the output layer\\
 \vspace{-0.85cm}
   \item[$z_{j}^{l}$] The pre-activation (linear component) term at neuron $j$ of layer $l$ in the IQNN\\
 \vspace{-0.85cm}
   \item[$a_{j}^{l}$] The activation term at neuron $j$ of layer $l$ in the IQNN\\
 \vspace{-0.85cm}
   \item[$\delta_{j}^{l}$] Binary variable represents the activation status at neuron $j$ of layer $l$ in the IQNN
 \end{itemize}
 
Using the above notation, the optimization problem can be formulated as follows:

\begin{subequations}\label{P1}
    \begin{align}
    \begin{split}
    \min_{x_s,n_s,\mathcal{X}} \quad &\sum_{s=1}^S\sum_{f=1}^F c^{L}u_{sf}d_{sf}\\
    &-8760 \times  \lambda^{\rm{PPA}} \left [ \frac{1}{1-\alpha}\sum_{o|\tau_{o}<1-\alpha} \left ( \frac{Q_{o}+Q_{o+1}}{2} \right ) \left ( \tau_{o+1}-\tau_{o} \right ) \right ] 
    \end{split}\label{eq:OF}\\
    &\text{s.t.}   \notag\\
    \label{P1_C1}&\qquad \sum_{s=1}^S n_{s} = \overline{N}
    \\
    \label{P1_C2}&\qquad \sum_{f=1}^F u_{sf} = 1,\qquad  s\in \{1,\dots,S\}\\
    \label{P1_C3}&\qquad u_{sf} \in \left \{  0,1\right \}, \qquad s\in \{1,\dots,S\}, f \in \{1,\dots,F\}\\
     \label{P1_C5}&\qquad a_{j}^{0}=\frac{\textbf{X}-\textbf{X}_{\min}}{\textbf{X}_{\max}-\textbf{X}_{\min}}, \qquad j \in \{1,...,n^{0}\}\\
     \label{P1_C6}&\qquad Q_{1}=a_{1}^{L}\left ( \textbf{Y}_{\max} -\textbf{Y}_{\min} \right ) +\textbf{Y}_{\min}\\ 
     \label{P1_C7}&\qquad Q_{o}=q_{o-1}+ a_{o}^{L}\left ( \textbf{Y}_{\max} -\textbf{Y}_{\min} \right ), \qquad o \in \{1,...,n^{L+1}=O\}\\  
     \label{P1_C8}&\qquad z_{j}^{l}=\sum_{i=1}^{n_{l}}w^{l}_{ij}a^{l-1}_{j}+b^{l}_{j}, \qquad  j \in \{1,...,n^{l}\}, l \in \{1,...,L\}\\
     \label{P1_C9}&\qquad z_{j}^{l} \leq a_{j}^{l}, \qquad  j \in \{1,...,n^{l}\}, l \in \{1,...,L\}\\ 
     \label{P1_C10}&\qquad a_{j}^{l} \leq z_{j}^{l}-M^{-,l}_{j}\left ( 1-\delta_{j}^{l}\right ) , \qquad  j \in \{1,...,n^{l}\}, l \in \{1,...,L\}\\ 
     \label{P1_C11}&\qquad a_{j}^{l} \leq M^{+,l}_{j}\delta_{j}^{l} , \qquad  j \in \{1,...,n^{l}\}, l \in \{1,...,L\}\\ 
    \label{P1_C12}&\qquad \delta_{j}^{l} \in \left \{  0,1\right \},   \qquad j \in \{1,...,n^{l}\}, l \in \{1,...,L\}
    \end{align}
\end{subequations}where $\mathcal{X}=\{u_{sf},d_{sf},Q_{o},z_{j}^{l},a_{j}^{l},\delta_{j}^{l}\}$ and $\textbf{Y}=[Q_{1},\dots,Q_{O}]^{T}$ represents the quantiles of wind energy outputs.

In the objective function \eqref{eq:OF}, the first term represents annualized cost of transmission lines, and the second term approximates the CVaR of RES power output. This is computed as a weighted average of those wind power quantiles that are below a predefined confidence level $1-\alpha$. Lower values of $1-\alpha$ reproduce more risk averse behaviors as we account only for the worst possible, and less probable, wind power outputs. On the contrary, assuming $1-\alpha=1$ is equivalent to a risk-neutral case, as we include all the quantiles approximating the wind power expected value. Following a common practice nowadays, we assume that the wind power is remunerated at a fixed price during the target year ($\lambda^{\rm{PPA}}$), resulting from a Power Purchase Agreement (PPA).

Constraint \eqref{P1_C1} fix the total number of constructed turbines for the wind farm sites $s \in \{1,\dots,S\}$. Constraint \eqref{P1_C2} enforces that the wind farm $s$ is only connected to one substation $f$ through the use of a binary variable $u_{sf}$ \eqref{P1_C3}. Constraints \eqref{P1_C5}-\eqref{P1_C12} represent the trained IQNN CL model.

Problem \eqref{P1} can be viewed as a two-stage stochastic optimization problem. The first-stage decisions would be the location, size and connection of each wind farm $s$. The resulting power production is uncertain, and characterized by its quantiles $Q_o$, and can be viewed as second-stage decision variables as they are conditioned by the aforementioned first-stage decisions through the embedded CL model. 

\section{Computational experiments}\label{CR}

\subsection{Data and experimental setup}\label{Data generation}

\subsubsection{Data generation}\label{Inp&Outp}

\begin{figure}[!h] 
\centering 
\includegraphics[width=0.7\textwidth]{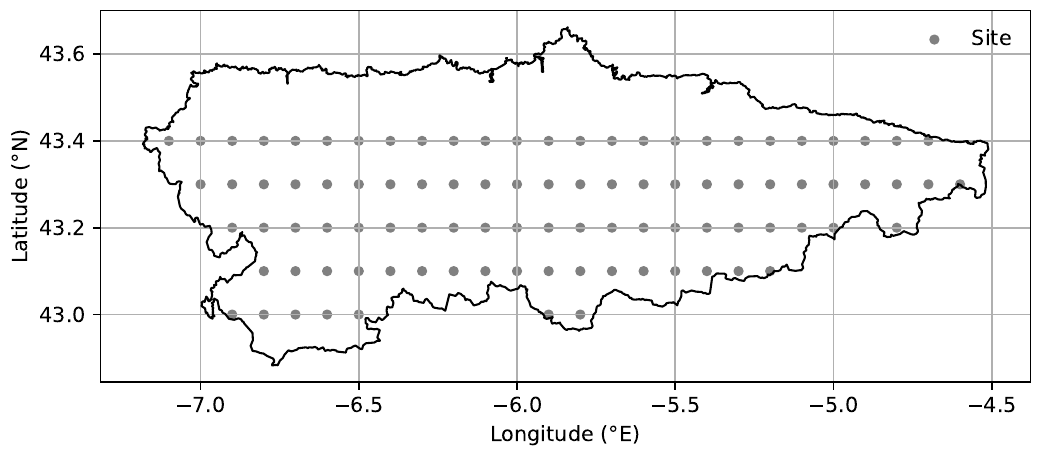} 
\caption{0.1\degree × 0.1\degree (latitude, longitude) grid covering Asturias} %
\label{Fig.WFA} %
\end{figure}

To train the IQNN (see Section \ref{sec_3_1}), we require a large set of realistic instances. In our case study, we consider two wind farms $S=2$ located in a non-coastal area of Asturias. We start considering a grid of coordinates with a spatial resolution of 0.1\degree × 0.1\degree (latitude, longitude), as shown in Figure \ref{Fig.WFA}. We then download the wind speed at each point of the grid, based on one year of ERA5-Land climate reanalysis data \citep{ERA5_2023}, covering the period from January 1, 2021, to December 31, 2021. Finally, the power production is calculated based on the power curve characteristics of the Vestas V90 2.0 MW turbine \citep{dhople2012framework}.

We generate training samples for two wind farms by running 1000 Monte Carlo iterations. In each iteration, two sites are randomly selected from the spatial grid, 3000 hourly timestamps are sampled from one year of hours, a total number of turbines is assigned randomly to the chosen sites, and the total hourly power production is computed. Finally, the generated samples are partitioned into training, validation, and test sets in proportions of 70\%, 15\%, and 15\%, respectively.

In order to embed our IQNN model into the optimization problem, the wind statistic are captured by the following terms:
\begin{itemize}
  \item Latitude of wind farm $s=1,2$
   \vspace{-0.3cm}
  \item Longitude of wind farm $s=1,2$
   \vspace{-0.3cm}
  \item Number of turbines of wind farm $s=1,2$
   \vspace{-0.3cm}
  \item 10m $v$ speed component of wind farm $s=1,2$: the northward component of the wind velocity measured at a height of 10 meters above the Earth's surface
   \vspace{-0.3cm}
  \item 10m $u$ speed component of wind farm $s=1,2$: the eastward component of the wind's horizontal speed at a height of 10 meters above the Earth's surface
   \vspace{-1cm}
  \item total wind production of wind farm $s=1$ and $s=2$
\end{itemize}

The optimization decision variables of siting (latitude and longitude) and sizing (number of turbines per site) in the optimization problem \eqref{P1} are considered as features for the IQNN training, whereas the total wind production, calculated from the wind-speed components, is considered as the target variable at this stage.

\subsubsection{Sample Analysis}
\begin{figure}[!h] 
\centering 
\includegraphics[width=1\textwidth]{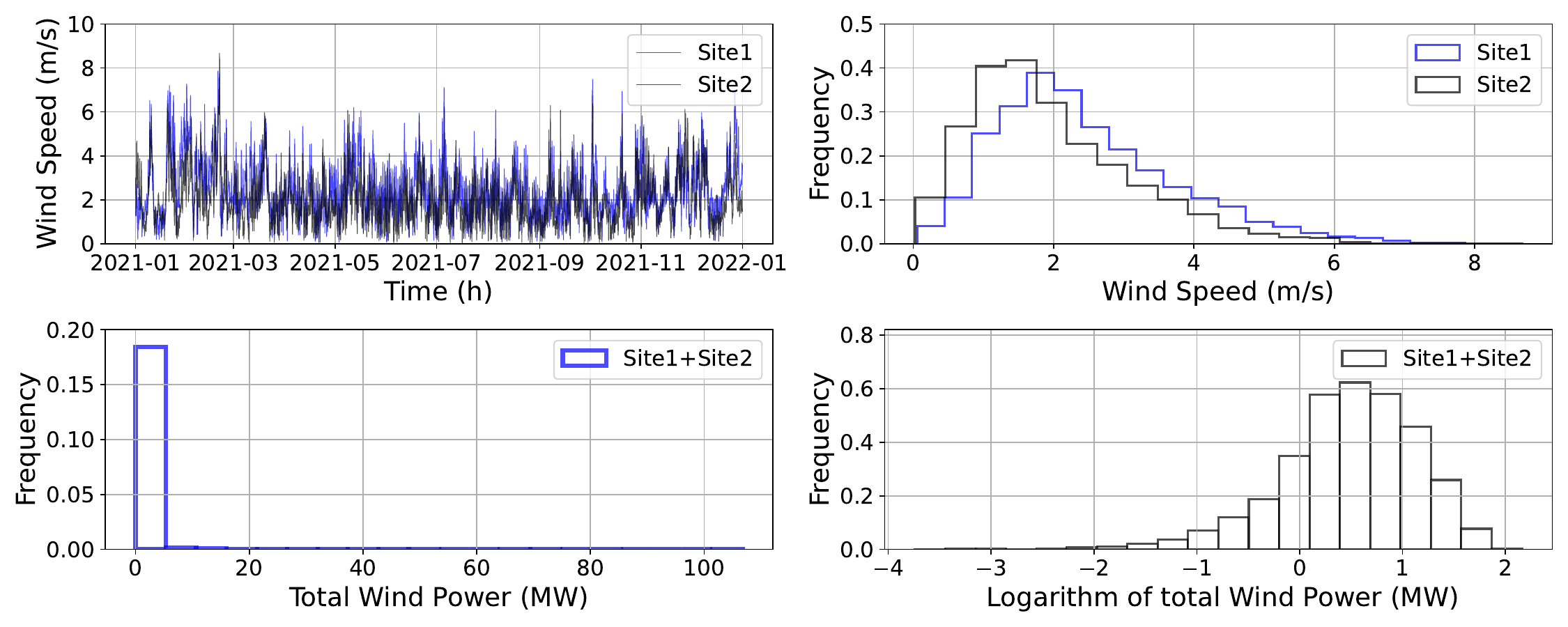} 
\caption{Wind speed and power curves for wind farms sample, with sitting coordinates of $\bigl\{(43.0 \degree N, -7.1 \degree E),(42.9 \degree N, -5.5 \degree E)\bigr\}$ and sizing of $(20,20)$ turbines} %
\label{Fig.PC} %
\end{figure}

Figure \ref{Fig.PC} presents the distribution of wind speed and total power production for two representative wind farm locations: Site 1 $(43.0 \degree N, -7.1 \degree E)$ and Site 2 $(42.9 \degree N, -5.5 \degree E)$. 
Wind speeds in Site 1 are significantly greater than in Site 2 as their time series and empirical distributions show.
The total wind energy output is computed by converting wind speed into power, by using the corresponding power curve for the considered wind turbine. Since the original power output values are highly skewed, a logarithm transformation\footnote{Power$_{log}$ = log(Power$_{raw}$+$\epsilon$), where $\epsilon>0$ is a small constant added to avoid taking the logarithm of zero.} is applied to rescale them into a more symmetrical distribution, which is better suited for IQNN model training. 

\subsubsection{Experimental setup}

The wind farm siting and sizing problem was solved by two steps. First, the IQNN learning model of wind power outputs was implemented in Python 3.11.9 using the Torch framework. Second, for the MILP formulation of the stochastic optimization model (\ref{P1}), we use Gurobi Optimizer version 11.0.1 under Python 3.11.9 on a system with an Intel i7 processor with 64GB RAM computer clocking at 2.90 GHz. We set a time limit of six hours (21,600 s) and the optimality gaps are relaxed to $1\%$.

\subsection{IQNN performance}

\subsubsection{Structure and quantile forecasts}

\begin{figure}[H] 
\centering 
\includegraphics[width=0.8\textwidth]{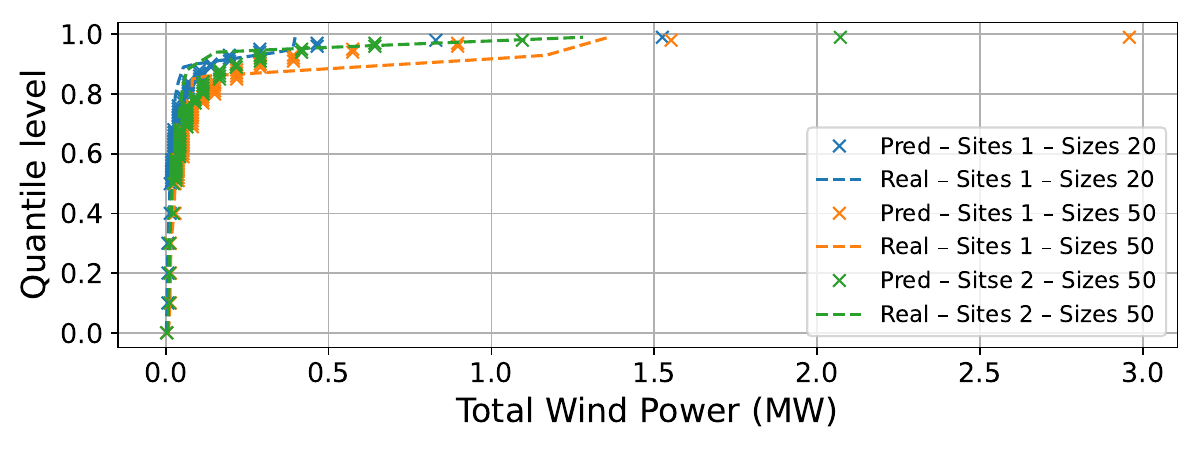} 
\caption{IQNN prediction and empirical real values of wind generation quantiles for three samples, including Sites 1 of $\bigl\{(42.9 \degree N, -7.1 \degree E),(43.2 \degree N,-6.4 \degree E)\bigr\}$, Sites 2 of $\bigl\{(43.1 \degree N, -6.8 \degree E),(42.9 \degree N, -6.3 \degree E)\bigr\}$. Each wind farm has the same turbine size of $(20,20)$ or $(50,50)$.} %
\label{Fig.QNNPre} %
\end{figure}

IQNN predicts total energy output at multi-quantile level. The structure of IQNN is instantiated as follows. The IQNN takes siting and sizing features from Section \ref{Inp&Outp} as input, then passes them to two hidden layers. Each hidden layer comprises 64 neurons. IQNN finally produces 55 quantiles of the wind power forecasts in its output layer. Specifically, multiple quantiles are arranged at 0.1-step intervals below the median quantile level, and 0.01-step intervals over the median quantile level, which capture the skewed wind generation profiles. Figure \ref{Fig.QNNPre} and Table \ref{table_EI} indicate our IQNN model approaches and capture the distribution of real total wind generation with averaged mean absolute error (Avg. MAE) of 0.11 (MW) and averaged root mean squared error (Avg. RMSE) of 0.27 (MW).

\begin{table}[!h]
\centering
\caption{Evaluation indexes of IQNN on sample datasets}
\label{table_EI}
\resizebox{10cm}{!}{
\begin{tabular*}{0.75\textwidth}{@{\extracolsep{\fill}}lcc}
    \toprule
    Data Set          &Avg. MAE (MW)        &Avg. RMSE (MW)       \\ 
    \midrule
    Validation Set & 0.104160335 & 0.236008374\\
    Test Set & 0.111868957 & 0.273837902 \\
    \bottomrule
\end{tabular*}}
\end{table}

\subsubsection{Comparison against bilinear interpolation}

\begin{figure}[H]
\centering
\begin{tikzpicture}[scale=1.0]
  \coordinate (A) at (0,0);
  \coordinate (B) at (3,0);
  \coordinate (C) at (0,3);
  \coordinate (D) at (3,3);

  \coordinate (P)  at (1.6,1.1);  
  \coordinate (Pc) at (1.5,1.5);  

  \draw[thick] (A) -- (B) -- (D) -- (C) -- cycle;

  \draw[dashed] (1.5,0) -- (1.5,3);
  \draw[dashed] (0,1.5) -- (3,1.5);

  \fill[blue]  (P)  circle (2pt) node[right] {$x_{s}=\,(lat,lon)$};
  \fill[black] (Pc) circle (2pt) node[above right] {center $=\,(lat_\ell,lon_k)$ };

  \draw[thick] (A) -- (B) -- (D) -- (C) -- cycle;

  \fill (A) circle (2.0pt) node[below left] {$(lat_{\ell-1},lon_{k-1})$};
  \fill (B) circle (2.0pt) node[below right] {$(lat_{\ell-1},lon_{k+1})$};
  \fill (C) circle (2.0pt) node[above left] {$(lat_{\ell+1},lon_{k-1})$};
  \fill (D) circle (2.0pt) node[above right] {$(lat_{\ell+1},lon_{k+1})$};

  \draw[->,gray] (A) -- (P);
  \draw[->,gray] (B) -- (P);
  \draw[->,gray] (C) -- (P);
  \draw[->,gray] (D) -- (P);
\end{tikzpicture}
\caption{Geometry instance of bilinear interpolation.}
\label{BI_interp}
\end{figure}
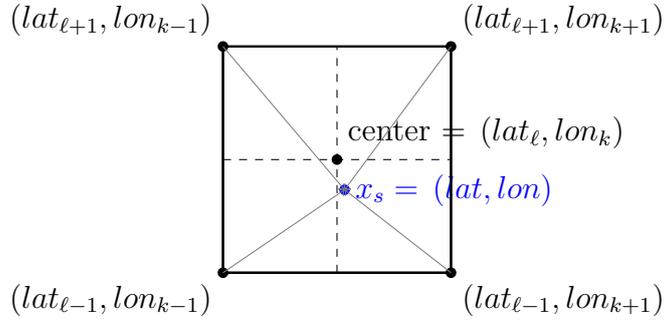

Once trained, the IQNN surrogate model for power quantiles allows evaluating any point within the considered region, extending the discrete geographical grid of the original dataset. To assess the effectiveness of IQNN for wind power quantiles, we set bilinear interpolation as the benchmark on an out-of-sample test set. Given grid unit corners $\{(lat_{\ell-1},lon_{k-1}),(lat_{\ell+1},lon_{k-1}),(lat_{\ell-1},lon_{k+1}),(lat_{\ell+1},lon_{k+1})\}$ with vertical and horizontal wind speed values $v_{\ell-1,k-1},\,v_{\ell+1,k-1},\,v_{\ell-1,k+1},\,v_{\ell+1,k+1}$ and $u_{\ell-1,k-1},\,u_{\ell+1,k-1},\,u_{\ell-1,k+1},\,u_{\ell+1,k+1}$, respectively, define the local coordinates  $x_s=(lat,lon)$ inside that unit, as depicted in Figure \ref{BI_interp} (blue dot).

The wind speed $f=\{u,v\}$ at coordinate $x_s$ is calculated based on the bilinear interpolation \citep{BOVIK200521,lowery2014optimizing}:
\begin{equation}
\label{}
f=(1-a)(1-b)f_{\ell-1,k-1}
+a(1-b)f_{\ell+1,k-1}
+(1-a)bf_{\ell-1,k+1}
+abf_{\ell+1,k+1}
\end{equation}
where parameters $a=\frac{lat-lat_{\ell-1}}{lat_{\ell+1}-lat_{\ell-1}}\in[0,1],b=\frac{lon-lon_{k-1}}{lon_{k+1}-lon_{k-1}}\in[0,1]$.

To perform this comparison, we consider the original ERA5 data set with a grid resolution of $0.1 \degree$, so wind information is available only at discrete latitude and longitude nodes. To evaluate estimating performance, we leave centered vertices $(lat_{\ell},lon_{k})$ out of the sample and train the IQNN. Then we forecast the power quantiles at those centered vertices by using the trained IQNN and the above bilinear interpolation method (see Figure \ref{BI_interp}). Specifically, we first interpolate the real $u$/$v$ wind speeds at each center from its four surrounding corners and then calculate wind generation quantiles. The empirical quantiles of out-of-sample vertices are used as the ground truth to compute the wind power quantile residuals for the two methods ($\epsilon_{IQNN, \tau}=\hat Q^{IQNN}_\tau-Q^{Emp}_\tau$ and $\epsilon_{Interp,\tau}=\hat Q^{Interp}_\tau-Q^{Emp}_\tau$).

\begin{figure}[!h] 
\centering 
\includegraphics[width=0.7\textwidth]{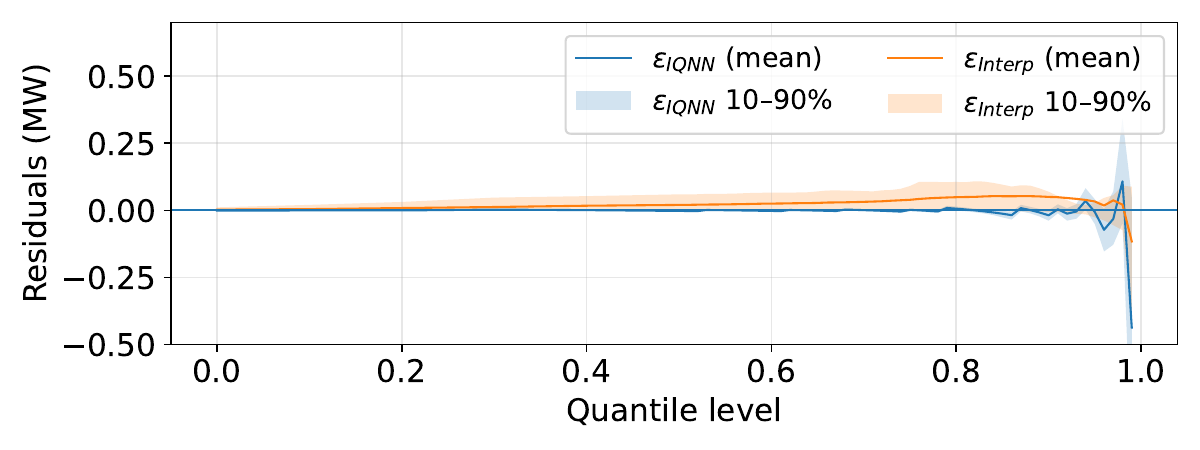} 
\caption{Predicted residuals comparison of IQNN ($\epsilon_{IQNN}$) and bilinear interpolation ($\epsilon_{Interp}$)} %
\label{Fig.IQNNvsInterp} %
\end{figure}

Figure \ref{Fig.IQNNvsInterp} shows that IQNN outperforms bilinear interpolation and yields small residuals at quantiles from 0.1 to 0.97. This surrogate model learns a piecewise-linear spatial response surface rather than discrete grid points. 
Because of these two merits, the IQNN surrogate model is well-suited to be used in the subsequent optimization process. 

\subsection{Case studies: risk-averse siting and sizing}
\subsubsection{Instance sites}
\label{sec:instance_sites}

\begin{figure}[!h]
\centering
\subfigure[Dominant generation within region]{%
    \label{RQ_a}
    \includegraphics[width=0.49\textwidth]{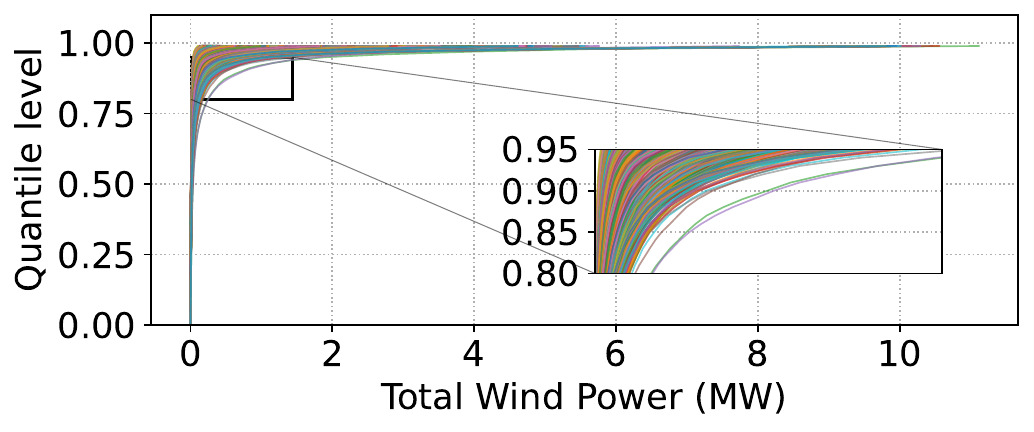}
}
\hspace{-5mm}
\subfigure[Disparity (differences) evaluation]{%
    \label{RQ_b}
    \includegraphics[width=0.49\textwidth]{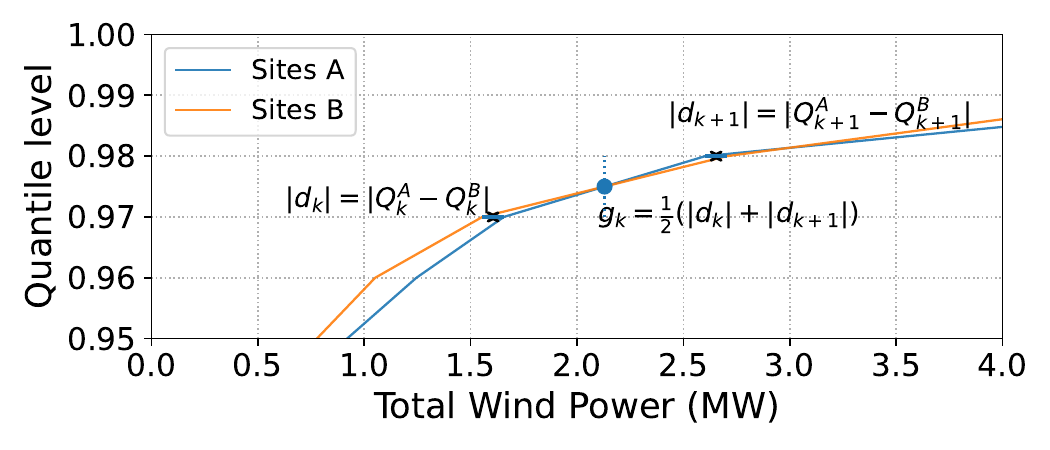}
}
\vspace{-2mm}
\subfigure[Clustering(K-means) on wind energy difference]{%
    \label{RQ_c}
    \includegraphics[width=0.49\textwidth]{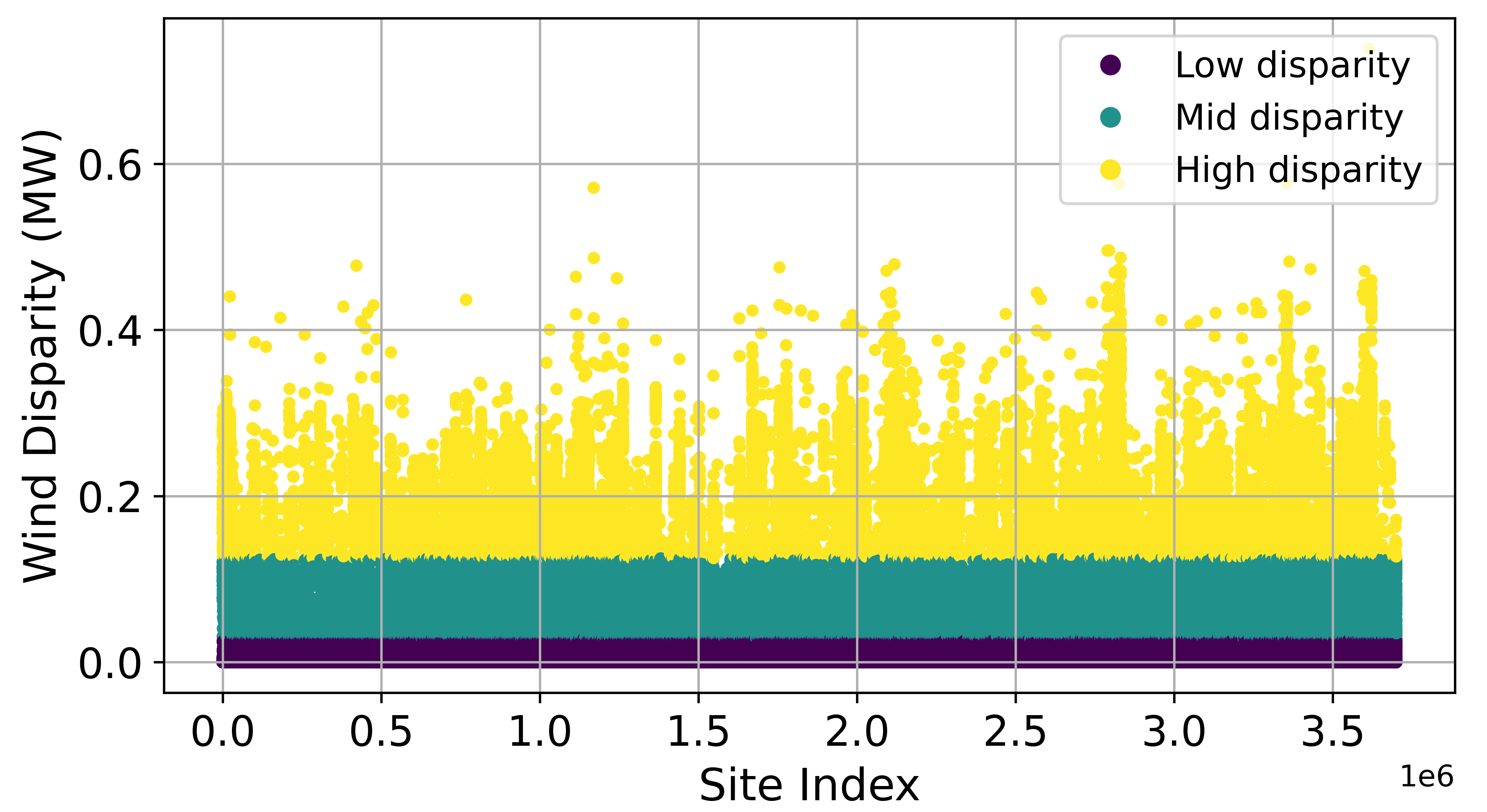}
}
\hspace{-5mm}
\subfigure[Subregion selection]{%
    \label{RQ_d}
    \includegraphics[width=0.49\textwidth]{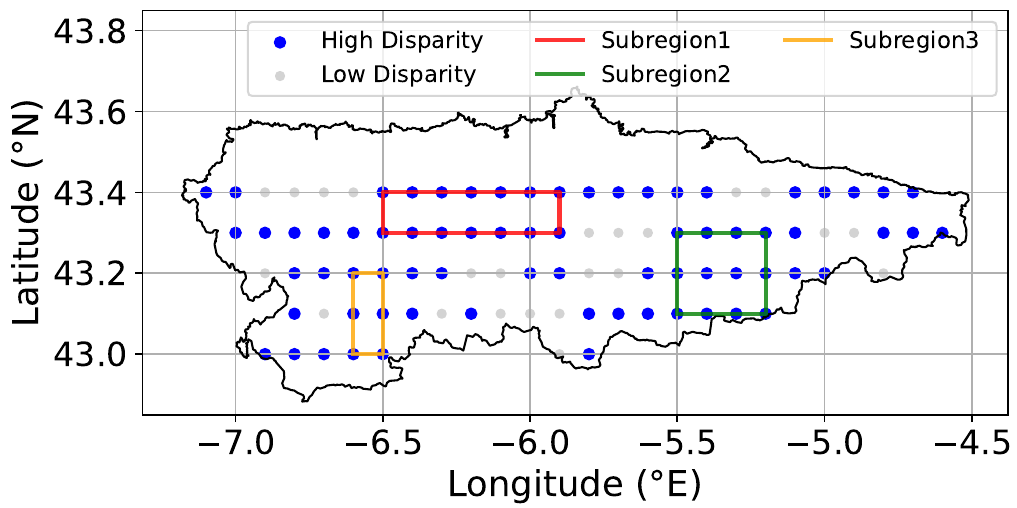}
}
\caption{Steps for instance site selection} 
\label{Fig.RQ} 
\end{figure}

Representative instance sites are selected to evaluate our proposed approach. 
Before we proceed with the analysis, we have identified that there exists a dominant phenomenon (see Figure \ref{RQ_a}), where certain locations consistently exhibit higher wind energy quantiles, for all confidence levels, than all other sites within the region. This results in optimal wind placements that do not change with the investor's risk-averse level. Assuming these locations might not be available, and for a more complete analysis, we have identified subregions where this phenomenon is not observed, i.e., there are non-dominant sites with crossing empirical distributions. In particular, optimizing wind placements in non-dominant sites is more challenging: no single portfolio of wind farms outperforms others for all confidence levels. Figure \ref{RQ_b} depicts two sites with cross-quantiles. The difference at intersected quantile ($k$) between sites A and B is denoted as $g_{k}$, which is calculated by horizontal gap of power quantiles $d_{k}$ and $d_{k+1}$. We define the total difference ($\sum_{k \in K} g_{k}$) between intersecting quantile curves as disparity.

We compute the disparity between intersecting curves for all non-dominant locations and then apply K-means clustering to categorize them based on their total disparity, as shown in Figure \ref{RQ_c}. The three high-disparity subregions selected through this process are illustrated in Figures \ref{RQ_d}. In the following case study, we implement wind farm placement optimization for both the entire region and these three high-disparity subregions.

\subsubsection{Results and analysis}

The following experiments examine risk-averse decisions for regional wind farm placement. The siting and sizing decisions are firstly investigated separately, including optimizing siting with fixed sizing, and optimizing sizing with fixed siting. Later, case studies on joint risk-averse optimization of siting and sizing are explored, explicitly incorporating transmission-connection costs. 

Regarding the economic and technical setup of the experiments, the power purchase agreement price for electricity, $\lambda^{PPA}$ is set as 60 \euro/MWh, while the fixed total number of wind turbines $\overline{N}$ is set as 40. The unit cost of transmission line is set as 5000 \euro/km. Finally, two substations are set at the locations (43.55\degree\,N, -5.79\degree\,E) and (43.30\degree\,N,  -6.67\degree\,E). 

\paragraph{\textbf{Risk-averse regional placement of fixed-size wind farms}}

We examine how the optimal placement evolves as the risk-aversion level varies for a fixed number of 20 turbines at each farm. Note that, the higher $1-\alpha$ stands for lower risk-aversion level. When $1-\alpha=1$, the CVaR$_\alpha$ is the expected value. Subregions (Region 1, 2, 3) are selected by removing the dominant coordinates from the full potential set (See Section \ref{sec:instance_sites}).

From Table \ref{FixedTurbine}, we observe that, as the risk aversion decreases (higher $1-\alpha$), the two wind farms locations converge to a single coordinate. The all-region solution is therefore located at (43.40\degree\,N, −7.10\degree\,E). Regarding the regional split, the unique coordinate for Region 1 is located at (43.40\degree\,N, −6.50\degree\,E), and for Region 3 is located at (43.20\degree\,N, −6.60\degree\,E). These results reflect that, as objectives increasingly aligned with the expected values, wind revenue dominance outweighs the diversification benefit. It leads to both farms optimally select the same location. In contrast, at the higher risk-aversion level $1-\alpha = 20\%$, the all-region solution places the farms at two distinct sites, and the resulting complementarity reduces the probability of simultaneous low production. 

\begin{table*}[h!]
\scriptsize
\centering
\caption{Optimize sites with fixed sizes $(20,20)$ under risk level $(1-\alpha)$}
\label{FixedTurbine}
\begin{tabularx}{\textwidth}{>{\raggedright\arraybackslash}c *{5}{>{\centering\arraybackslash}X}}
\toprule
\multirow{2}{*}{$1-\alpha$} &
\multicolumn{5}{c}{Optimal sites and CVaR of revenues (\euro) } \\
\cmidrule(lr){2-6}
& All regions & Regions 1–3 & Region 1 & Region 2 & Region 3 \\
\midrule
20\% &
\coords{43.40\degree\,N,\,-7.10\degree\,E}{43.40\degree\,N,\,-4.69\degree\,E}{8230.81} &
\coords{43.12\degree\,N,\,-6.60\degree\,E}{43.40\degree\,N,\,-5.90\degree\,E}{5090.04} &
\coords{43.31\degree\,N,\,-6.50\degree\,E}{43.40\degree\,N,\,-5.90\degree\,E}{4998.74} &
\coords{43.10\degree\,N,\,-5.20\degree\,E}{43.30\degree\,N,\,-5.23\degree\,E}{4728.05} &
\coords{43.20\degree\,N,\,-6.60\degree\,E}{43.14\degree\,N,\,-6.60\degree\,E}{5027.96} \\
\addlinespace[3pt]

57\% &
\coords{43.40\degree\,N,\,-7.10\degree\,E}{43.40\degree\,N,\,-4.69\degree\,E}{43124.80} &
\coords{43.34\degree\,N,\,-6.50\degree\,E}{43.40\degree\,N,\,-6.50\degree\,E}{16946.06} &
\coords{43.40\degree\,N,\,-6.50\degree\,E}{43.40\degree\,N,\,-6.50\degree\,E}{17015.90} &
\coords{43.30\degree\,N,\,-5.21\degree\,E}{43.30\degree\,N,\,-5.50\degree\,E}{13449.65} &
\coords{43.20\degree\,N,\,-6.55\degree\,E}{43.20\degree\,N,\,-6.60\degree\,E}{16298.80} \\
\addlinespace[3pt]

67\% &
\coords{43.40\degree\,N,\,-7.10\degree\,E}{43.40\degree\,N,\,-4.69\degree\,E}{58815.11} &
\coords{43.40\degree\,N,\,-6.50\degree\,E}{43.40\degree\,N,\,-6.50\degree\,E}{20610.42} &
\coords{43.40\degree\,N,\,-6.50\degree\,E}{43.40\degree\,N,\,-6.50\degree\,E}{20610.42} &
\coords{43.30\degree\,N,\,-5.50\degree\,E}{43.30\degree\,N,\,-5.20\degree\,E}{15728.21} &
\coords{43.18\degree\,N,\,-6.60\degree\,E}{43.20\degree\,N,\,-6.60\degree\,E}{19534.90} \\
\addlinespace[3pt]

77\% &
\coords{42.90\degree\,N,\,-7.10\degree\,E}{43.40\degree\,N,\,-7.10\degree\,E}{56662.88} &
\coords{43.40\degree\,N,\,-6.50\degree\,E}{43.40\degree\,N,\,-6.50\degree\,E}{26719.20} &
\coords{43.40\degree\,N,\,-6.50\degree\,E}{43.40\degree\,N,\,-6.50\degree\,E}{26719.20} &
\coords{43.30\degree\,N,\,-5.44\degree\,E}{43.30\degree\,N,\,-5.20\degree\,E}{19052.00} &
\coords{43.20\degree\,N,\,-6.60\degree\,E}{43.20\degree\,N,\,-6.60\degree\,E}{24788.19} \\
\addlinespace[3pt]

87\% &
\coords{43.40\degree\,N,\,-7.10\degree\,E}{43.40\degree\,N,\,-7.10\degree\,E}{119571.21} &
\coords{43.40\degree\,N,\,-6.50\degree\,E}{43.30\degree\,N,\,-6.50\degree\,E}{38152.34} &
\coords{43.39\degree\,N,\,-6.50\degree\,E}{43.40\degree\,N,\,-6.50\degree\,E}{38033.16} &
\coords{43.30\degree\,N,\,-5.50\degree\,E}{43.30\degree\,N,\,-5.50\degree\,E}{24529.45} &
\coords{43.20\degree\,N,\,-6.60\degree\,E}{43.20\degree\,N,\,-6.60\degree\,E}{33932.20} \\
\addlinespace[3pt]

97\% &
\coords{43.40\degree\,N,\,-7.10\degree\,E}{43.40\degree\,N,\,-7.10\degree\,E}{314840.84} &
\coords{43.40\degree\,N,\,-6.50\degree\,E}{43.30\degree\,N,\,-6.50\degree\,E}{72892.30} &
\coords{43.40\degree\,N,\,-6.50\degree\,E}{43.40\degree\,N,\,-6.50\degree\,E}{72892.30} &
\coords{43.30\degree\,N,\,-5.50\degree\,E}{43.30\degree\,N,\,-5.20\degree\,E}{37091.90} &
\coords{43.20\degree\,N,\,-6.60\degree\,E}{43.20\degree\,N,\,-6.60\degree\,E}{60393.45} \\
\bottomrule
\end{tabularx}
\end{table*}

\begin{figure}[h!]
\centering
\subfigure[Quantiles variation, All regions]{%
    \label{All_quantile}
    \includegraphics[width=0.49\textwidth]{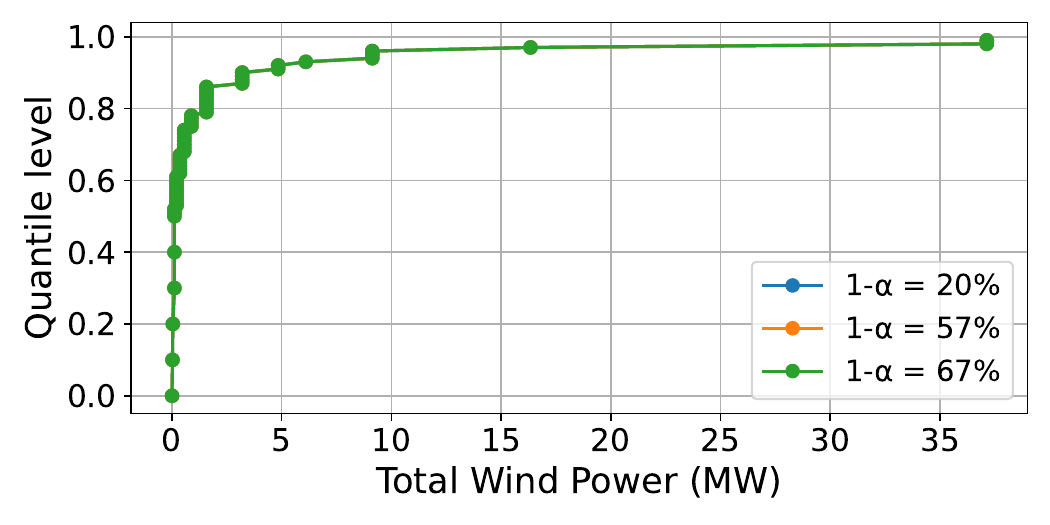}
}
\hspace{-5mm}
\subfigure[Location variation, All regions]{%
    \label{All_sites}
    \includegraphics[width=0.49\textwidth]{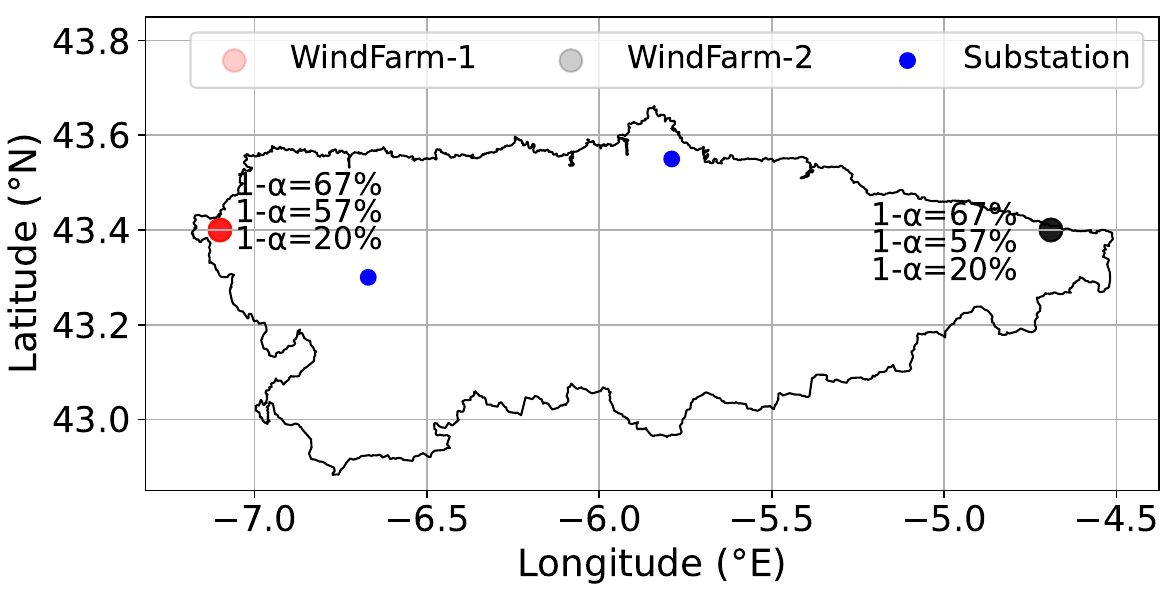}
}
\vspace{-2mm}
\subfigure[Quantiles variation, Region 1-3]{%
    \label{Region123_quantile}
    \includegraphics[width=0.49\textwidth]{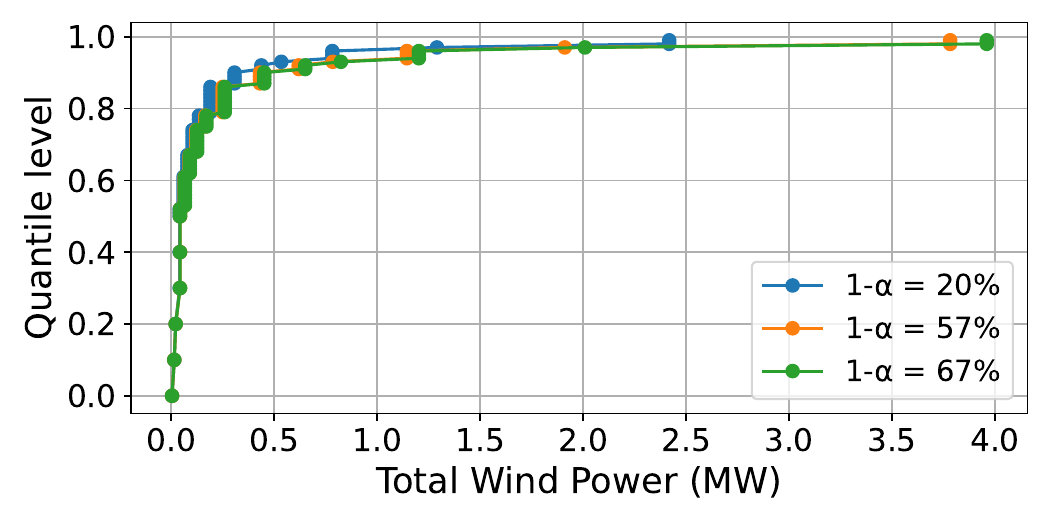}
}
\hspace{-5mm}
\subfigure[Location variation, Region 1-3]{%
    \label{Region123_sites}
    \includegraphics[width=0.49\textwidth]{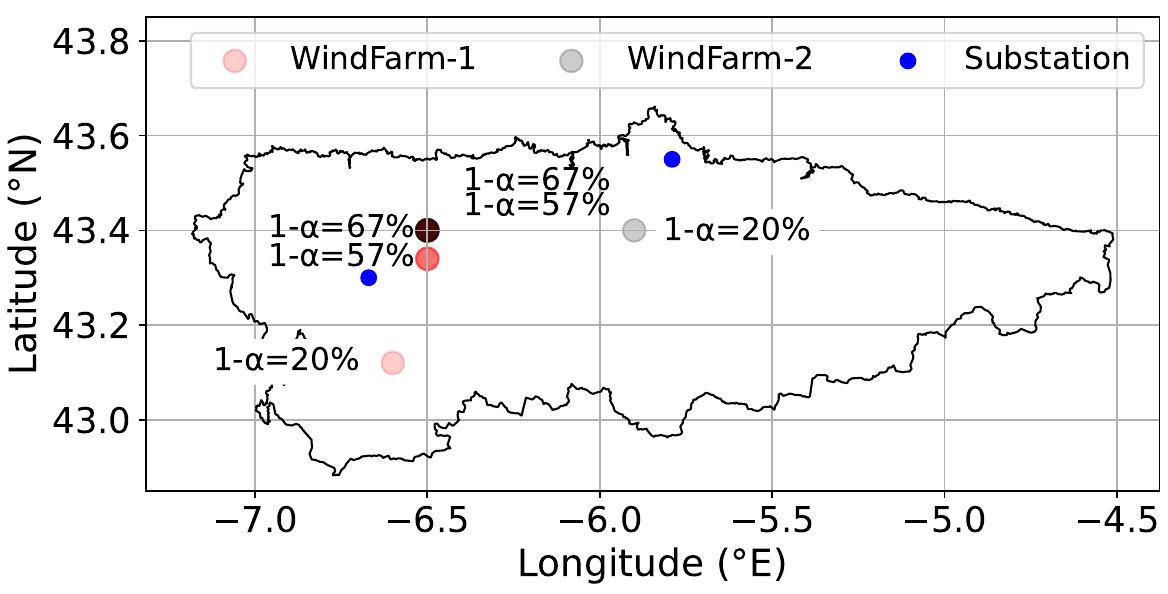}
}
\caption{Wind farm placement under risk aversion, in the entire region and high-disparity subregion 1} 
\label{Fig.RA} 
\end{figure}

Graphically, Figure \ref{Fig.RA} shows the detailed wind farm placement in dominant (All regions) and non-dominant regions (joint Region 1–3). We can notice joint Region 1–3 selects sites that span both Region 1 and Region 2 at higher risk-aversion level $1-\alpha = 20\%$ (see Figure \ref{Region123_sites}), illustrating cross-regional spatial diversification. Overall, the results indicate that wind farm placement benefits from spatial diversification when the risk-aversion level is high. It is because CVaR targets the extremely worst but stable revenues, and thereby favors widely separated sites. In contrast, as the level of risk aversion decreases ($1-\alpha$ increases), the CVaR converges to the expected value and concentrates the wind turbines in the dominant locations for high wind generation.

\paragraph{\textbf{Risk-averse size portfolios of fixed-placement wind farms}}

Given the fixed siting and a total of 40 turbines, optimal size portfolios spread sensitively at each risk aversion level. Table \ref{FixedSite} shows that, in cases of All region and Region 1-3, the dominant site retains most of the turbines and the size portfolio remains stable as risk aversion decreases (higher $1- \alpha$). The Region 1 results reveal sensitivity to marginal revenue. When the two fixed sites have nearly equal incremental revenue per turbine, the optimizer exhibits an abrupt sizing switch (from (32,8) to (19,21) between 57\% and 67\%). 

Moreover, we can observe a complete shift on the size decisions for Region 2, going from (0,40) to (40,0) as the risk-aversion decreases. This may indicate that wind power quantiles of wind farm 1 are dominant at a high risk-aversion level, while wind power quantiles of wind farm 2 are dominant at a low risk-aversion levels. In Region 3, on the contrary, the turbine allocation becomes more balanced as the risk-aversion decreases ($1-\alpha$ increases). For instance, turbines transition from (0,40) at 20\% to approximately (17,23) at 97\%. These results suggest that the impact of the risk-aversion level on the optimal size of each wind farm is highly sensitive to the particular region under study. 

\begin{table*}[h!]
\begingroup
\scriptsize       
\centering
\caption{Optimize sizes with fixed sites under risk level $(1-\alpha)$}
\label{FixedSite}
\begin{tabularx}{\textwidth}{>{\raggedright\arraybackslash}c *{6}{>{\centering\arraybackslash}X}}
\toprule
\multicolumn{7}{c}{Optimal sizes and CVaR of revenues (\euro) }\\
\cmidrule(lr){1-7}
$1-\alpha$ & 20\% & 57\% & 67\% & 77\% & 87\% & 97\% \\
\midrule
\coordsIII{All regions}{43.40\degree\,N,\,-7.10\degree\,E}{43.40\degree\,N,\,-4.70\degree\,E}&
\coordsII{30,10}{7540.59 } &
\coordsII{30,10}{35882.64 } &
\coordsII{30,10}{52158.02 } &
\coordsII{30,10}{82353.56 } &
\coordsII{30,10}{145546.77 } &
\coordsII{30,10}{378789.20 } \\[3pt]

\coordsIII{Regions 1-3}{43.10\degree\,N,\,-6.60\degree\,E}{43.40\degree\,N,\,-5.90\degree\,E} &
\coordsII{10,30}{4802.83 } &
\coordsII{8,32}{13659.52 }  &
\coordsII{8,32}{16324.03 } &
\coordsII{8,32}{20357.58 } &
\coordsII{11,29}{26939.47 }&
\coordsII{11,29}{44722.86 }\\[3pt]

\coordsIII{Regions 1}{43.30\degree\,N,\,-6.50\degree\,E}{43.40\degree\,N,\,-5.90\degree\,E} &
\coordsII{18,22}{4528.95} &
\coordsII{32,8}{12826.62}  &
\coordsII{19,21}{15584.09} &
\coordsII{19,21}{19700.02} &
\coordsII{18,22}{26874.65}&
\coordsII{19,21}{46716.01} \\[3pt]

\coordsIII{Regions 2}{43.10\degree\,N,\,-5.20\degree\,E}{43.30\degree\,N,\,-5.20\degree\,E} &
\coordsII{0,40}{5336.31} &
\coordsII{0,40}{12693.20}  &
\coordsII{0,40}{15019.32} &
\coordsII{0,40}{17993.06} &
\coordsII{37,3}{22974.04}&
\coordsII{40,0}{35018.37} \\[3pt]

\coordsIII{Regions 3}{43.20\degree\,N,\,-6.60\degree\,E}{43.10\degree\,N,\,-6.60\degree\,E} &
\coordsII{0,40}{4686.42} &
\coordsII{21,19}{11451.76} &
\coordsII{21,19}{14046.92} &
\coordsII{21,19}{17847.17} &
\coordsII{16,24}{24572.09}&
\coordsII{17,23}{46403.11} \\

\bottomrule
\end{tabularx}
\endgroup
\end{table*}

\paragraph{\textbf{Risk-averse joint placement and sizing with transmission connection costs}} 

Table \ref{TransmLine} presents the investment results for the All regions case when placement and sizing decisions are taken considering risk-aversion and transmission connection costs. Again, we consider a total of 40 turbines that need to be divided between the two wind farms. At low $1-\alpha$ (high risk aversion), the optimal sites are located near substations with zero connection distance (see Table \ref{TransmLine}). While as $1-\alpha$ increases and the objective shifts toward expected profits, the solution implies longer lines to access stronger average winds. The corresponding connection distance ranges from 0 km up to roughly 48 km at $1-\alpha = 77\%$ and about 133 km at $1-\alpha = 97\%$.

Table \ref{TransmLine} indicates that when the objective emphasizes lower‑tail outcomes, transmission cost strongly constrains siting. This is because tail wind quantiles are small, so the marginal gain of CVaR from relocating to remote, windier sites is limited by the incremental cost of longer lines. It indicates that increased weight on extremely low generation scenarios dampens incentives to build costly connections to distant substations. As $1-\alpha$ increases, the objective increasingly rewards high generation performance, and the optimal solution adopts longer connections to capture higher mean wind speeds. It consequently increases revenues CVaR, which covers higher line costs. This pattern is aligned with co‑optimization studies where lower effective risk aversion shifts investments toward high generation sites that improve average returns even at greater distance. These results provide insight on risk‑sensitive wind farm placement. Adding weight to tail outcomes (high risk aversion) tends to favor nearby, lower‑cost connections. However, reducing the weight on tails (larger $1-\alpha$) supports longer connections to windy sites and adapts farm sizes to exploit greater wind generation.

\begin{table*}[h!]
\begingroup
\scriptsize       
\centering
\caption{Jointly optimize sites and sizes over all regions with transmission cost under risk level $(1-\alpha)$}
\label{TransmLine}

\begin{tabularx}{\textwidth}{@{}%
  >{\centering\arraybackslash}p{1.2cm}  
  W{10}                                
  p{1.5cm}                              
  W{12} 
  >{\centering\arraybackslash}p{2.3cm}
  >{\centering\arraybackslash}W{8}  
@{}}
\toprule
\multirow{2}{*}{$1-\alpha$} &
\multicolumn{5}{c}{Optimal sites, sizes, cost and CVaR of revenues} \\
\cmidrule(lr){2-6}
& Siting & Sizing & Transmission distance (km) & Transmission cost (\euro) &CVaR of revenues (\euro)  \\
\midrule
20\%  &
\coordsI{43.30\degree\,N,\,-6.67\degree\,E}{43.30\degree\,N,\,-6.67\degree\,E} & (19,21) & \coordsI{$u_{1,2}=1, d_{1,2}=0$ }{$u_{2,2}=1,d_{2,2}=0$ } & 0  & 8404.46   \\
\addlinespace[3pt]

57\% & \coordsI{43.30\degree\,N,\,-6.67\degree\,E}{43.30\degree\,N,\,-6.67\degree\,E} & (19,21) & \coordsI{$u_{1,2}=1, d_{1,2}=0$ }{$u_{2,2}=1,d_{2,2}=0$ } & 0  & 16479.18   \\
\addlinespace[3pt]

67\% &\coordsI{43.40\degree\,N,\,-6.67\degree\,E}{43.30\degree\,N,\,-6.67\degree\,E} & (19,21) & \coordsI{$u_{1,2}=1, d_{1,2}=0$ }{$u_{2,2}=1,d_{2,2}=0$ } & 0  & 20190.06   \\
\addlinespace[3pt]

77\% &\coordsI{43.30\degree\,N,\,-6.67\degree\,E}{43.40\degree\,N,\,-7.10\degree\,E} & (8,32) & \coordsI{$u_{1,2}=1, d_{1,2}=0$ }{$u_{2,2}=1,d_{2,2}=48.50$ } & 15383.98  & 30025.82   \\
\addlinespace[3pt]

87\% &\coordsI{43.40\degree\,N,\,-7.10\degree\,E}{43.40\degree\,N,\,-7.10\degree\,E} & (25,15) &\coordsI{$u_{1,2}=1, d_{1,2}=48.50$ }{$u_{2,2}=1,d_{2,2}=48.50$ } &30767.96  & 63936.01 \\
\addlinespace[3pt]

97\% &\coordsI{43.40\degree\,N,\,-4.60\degree\,E}{43.40\degree\,N,\,-7.10\degree\,E} & (17,23) &\coordsI{$u_{1,1}=1, d_{1,1}=132.56$ }{$u_{2,2}=1,d_{2,2}=48.50$ } & 57436.23  &212413.17   \\
\bottomrule
\end{tabularx}
\endgroup
\end{table*}

\section{Conclusions and future research}\label{Conclusion}

Previous approaches optimize onshore wind farm placement on a finite set of sites. We propose a new data-driven approach for the optimal regional wind farm sitting and sizing which spatio-temporally extrapolates wind profiles across an entire region. We show its adequate performance on inferring the empirical distribution of hourly wind power production, given a set of wind farm sites. 

To this end, we employ a probabilistic incremental quantile neural network (IQNN) trained on real high-resolution regional data from a northern region in Spain. The IQNN includes ReLU activation functions, so that it can be rewritten as an equivalent mixed-integer linear set of constraints. We include these constraints into the stochastic sitting and sizing decision problem. 

Hence, the decision-making model implicitly accounts for the probabilistic impact that a set of sites has on the aggregate wind power production. This model can incorporate different risk-aversion levels by approximating the CVaR, making it suitable for risk-averse investors. Results suggest that, for fixed farm sizes, high risk aversion leads to spatial diversification, i.e. more distant wind farm locations. However, for the case of optimizing sizes with fixed sites we observe that the solution is highly sensitive to the behavior of wind power quantiles, which may vary across regions. Moreover, incorporating transmission connection cost affects significantly to the siting decisions. Risk-averse investor avoid long connecting lines, despite selecting lower production sites. On the contrary, risk-neutral investors are willing to move further from substations to achieve a higher average wind generation.

The proposed optimization model facilitates the portfolio of regional wind farms, and gives support for risk-averse investor’s decision making. Future research may consider a more detailed description of the candidate sites within a region with a richer set of locational constraints.

\section*{Acknowledgements} 
This work is part of the R\&D project PID2023-151013NB-I00, funded by MCIN/AEI/10.13039/501100011033 and by ERDF/EU. And Wenxiu Feng also acknowledges the support from China Scholarship Council, China, Grant No. 202106890026.

\bibliographystyle{elsarticle-harv}
\singlespacing
\fontsize{9.25}{10.25}\selectfont
\bibliography{referenceArticle}

\end{document}